# Verbalizing Ontologies in Controlled Baltic Languages

Normunds GRŪZĪTIS, Gunta NEŠPORE and Baiba SAULĪTE
*Institute of Mathematics and Computer Science, University of Latvia*

**Abstract.** Controlled natural languages (mostly English-based) recently have emerged as seemingly informal supplementary means for OWL ontology authoring, if compared to the formal notations that are used by professional knowledge engineers. In this paper we present by examples controlled Latvian language that has been designed to be compliant with the state of the art Attempto Controlled English. We also discuss relation with controlled Lithuanian language that is being designed in parallel.

**Keywords.** Controlled Natural Language, Ontology Verbalization, Information Structure, Synthetic Language, Baltic Languages

## Introduction

One of the fundamental requirements in verbalization of ontology structure, restrictions and data integrity constraints [1] is unambiguous interpretation of controlled natural language (CNL) statements, so that the CNL user could easily predict the precise meaning of the specification he/she is writing or reading. In the case of integrity constraints, the interpretation process also includes resolving of anaphoric references.

Several restrictions are used in CNLs to enable the deterministic construction of discourse representation structures (DRS): a strict syntactic subset of natural language, a set of interpretation rules for potentially ambiguous constructions, a monosemous (domain-specific) lexicon, an assumption that the antecedent of a definite noun phrase (NP) is the most recent and most specific accessible NP.

There are several sophisticated CNLs that provide seemingly informal means for bidirectional mapping between controlled English and OWL [1]. Although the existing CNLs primarily focus on English, Angelov and Ranta [2] have shown that the Grammatical Framework (GF), a formalism for implementing multilingual CNLs, provides convenient means for writing parallel grammars that simultaneously cover similar syntactic fragments of several natural languages. Thus, if the abstract and concrete grammars are carefully designed, GF provides syntactically and semantically precise translation from one CNL to another. This potentially allows exploitation of powerful tools that are already developed for controlled English also for non-English CNLs. For instance, the Attempto Controlled English (ACE) parser [3] could be used for DRS construction, paraphrasing and mapping to OWL, and ACE verbalizer [4] could be used in the reverse direction, facilitating cross-lingual ontology development, verbalization, and querying.

---

[1] Here we refer to OWL 2 terminological statements (http://www.w3.org/TR/owl2-primer/), SWRL implication rules (http://www.w3.org/Submission/SWRL/) and SPARQL integrity queries (http://www.w3.org/TR/rdf-sparql-query/).

While it seems promising and straightforward for rather analytical languages that share common fundamental characteristics, allowing (apart from other) explicit detection of given and new information and, thus, detection of anaphoric references, it raises issues in the case of highly synthetic languages (like Baltic languages), where explicit linguistic markers, indicating which information is new (potential antecedents) and which is already given (anaphors), in general, are not available. In analytical CNLs, analysis of the information structure of a sentence is based on the strict word order and systematic use of definite and indefinite articles. In highly synthetic languages, articles are rarely used[2] and are "compensated" by more implicit linguistic markers; typically, by changes in the neutral word order, which is enabled by rich inflectional paradigms and syntactic agreement.

We might impose the consistent use of artificial determiners, using, for example, indefinite and demonstrative pronouns, but then the controlled language would lose its characteristics of naturalness and intuitiveness. The problem is even more apparent in case of Lithuanian that, in contrast to Latvian, has not been historically influenced by German. Therefore the only[3] formal and general feature that indicates the status of a NP is its position in a sentence — whether it belongs to the topic or focus part. Thus, the correspondence between the given/new information and the word order patterns can be described in terms of topic-focus articulation (TFA) [5], i.e., what we are talking about and what we are saying new about it.

Although the topic (theme) and focus (rheme) parts of a sentence, in general, are not always reflected by systematic changes in the word order, it has been shown [6] that, in the case of controlled Latvian, TFA is a simple and reliable mechanism for deterministic (predictable) analysis of the information structure of a sentence. As the initial evaluation shows, the "correct" word order is both intuitively satisfiable by a native speaker and facilitates the automatic detection of anaphoric NPs in highly synthetic CNL.

In order to evaluate the TFA-based approach, an on-line questionnaire was created for each language, containing 15–17 ontological statements and rules of different complexity, each of them verbalized in two or three slightly different ways. The survey was aimed at a rather wide target group: ca. 80 respondents participated in the Latvian survey, and ca. 40 respondents participated in the Lithuanian survey (in both cases ca. 75% evaluated all examples; others — at least one third). Each of the proposed alternative verbalizations had to be ranked being either "good", "acceptable" or "poor". In addition, respondents were able to propose their own (modified) verbalizations — this option was frequently used leading to many interesting and/or overlapping suggestions.

In this paper we present by examples the improved and extended version of controlled Latvian language, if compared to [6], and discuss relation with controlled Lithuanian language that is being designed in parallel. At the end we briefly summarize some remaining issues and future tasks. Prototype implementations for both languages are available on-line at *http://eksperimenti.ailab.lv/cnl*.

---

[2] In Baltic languages, the (in-)definiteness feature is not encoded even in noun endings as it is in the case of Scandinavian languages, for instance.

[3] Although definite and indefinite adjectives and participles can be used in multi-word units, such markers are optional and, in the case of controlled language, non-reliable — attributes in domain-specific terms often have definite endings by default.

## 1. Generalization and SVO Statements

The most simple and basic type of ontological statements are generalizations defining that one named class is a subclass of another named class, i.e., statements that define the taxonomy of an ontology (see Table 1). But even in this simple case there is no consensus among respondents, which verbalization is the best, except the common agreement that the indefinite pronoun ("article") is absolutely unnecessary in predicate nominal phrases.

Apart from some nuances, the main dissatisfaction was about the singular subjects. Due to the underlying formalism — a subset of first-order logic (FOL) — in ACE and in other CNLs for OWL, references in the singular are typically used [1]. In natural language, however, plural statements are more intuitively and frequently used when generalizations are expressed. Thus, we have allowed in our grammar plural clauses in parallel to the singular ones: they are automatically paraphrased (linearized) into the singular readings, ensuring also compliance with ACE (see *Sg* vs. *Pl* in Table 1 and henceforth). Several respondents noted that they also would prefer to skip the plural determiner (universal quantifier) *visi* ('all'). We could allow this if we were interested only in terminological (TBox) statements, however, our aim is to cover rules (see Table 7 in the next section) and assertional statements as well, and therefore the optional determiner would introduce ambiguity (universal vs. existential quantification).

Paraphrasing mechanism in CNLs is often used to explicitly indicate the interpretation of a potentially ambiguous syntactic construction. Grammatical Framework is especially handy for dealing with paraphrases — we use them widely throughout the grammar; moreover, GF supports using this mechanism even at the lexical level, allowing synonyms for both function and content words (in Table 1 and henceforth synonym lists are given in parenthesis; the first word is used in linearization). For instance, the majority of Latvian respondents suggested the pronoun *ikviens* for 'every' instead of *katrs* — if it is used as an attribute of a noun.

As the survey showed, it should be emphasized that we are not addressing the machine translation problem in the traditional sense: the semantically precise translation (via OWL as interlingua) among the controlled languages is not the ultimate goal but rather a side-effect. Some statements in CNL might not conform to a fully correct subset of natural language — different trade-offs have to be made to ensure the predictable interpretation.

One of the main reasons for using a CNL is to encourage the active involvement of domain experts in the conceptual modeling phase of an ontology [7]. CNL provides high-level intuitive means for ontology authoring, if compared to the formal notations[4] that are used by professional knowledge engineers. The consequence of involving domain experts is that the conceptual ontology most likely will not be "optimally" and/or completely defined. Thus, the subsequent ontology modeling phase, involving a knowledge engineer, is needed in general. For example, it might be easier for a domain expert to define several subclasses of the same class in one sentence (see Table 2). To avoid the anonymous class the knowledge engineer might decide to split the axiom in two or more separate axioms. Moreover, the actually intended generalization in some cases might be in the inverse direction (as in Table 2).

**Table 1.** Generalization axiom from sample university ontology. *P* stands for parsing, *L* — for linearization. For the same linearization the parser can accept grammatically and/or lexically different statements.

| ACE | | **Every** professor **is a** teacher. |
|---|---|---|
| Sg | P | (Ikviens\|Katrs) profesors ir pasniedzējs. |
|    | L | **Ikviens** profesors **ir** pasniedzējs. |
| Pl | P | Visi profesori ir (pasniedzēji\|skolotāji). |
|    | L | **Visi** profesori **ir** pasniedzēji. |
| OWL | | Class: Professor SubClassOf: Teacher |

Another interesting point that was confirmed regarding the predicate nominals is that CNL users often would like to use the 'either .. or' (*vai nu .. vai*) construction instead of just 'or', although the intended meaning is the simple disjunction (*OR*) instead of the exclusive disjunction (*XOR*). Thus, we have allowed to use it while defining an axiom, but in the paraphrases the element 'either' is automatically dropped, indicating that the interpretation, perhaps, is not what was expected. If the exclusive disjunction was actually intended, it might be the case that additional disjointness axioms have to be defined (as illustrated in Section 3).

Subclass axioms often include property restrictions, making them more complex and, if verbalized in CNL, the generalization might not be explicitly visible (compare the verbalizations in Table 3 and Table 4). In terms of CNL, a property restriction is a subject-verb-object (SVO) statement (clause), where either the subject or the object is existentially quantified (in controlled English, it is always the object). Thus, the question about using the indefinite marker (pronoun) rises again[5]. In Lithuanian this would be ungrammatical, but in Latvian such markers might improve the reading in certain cases: the survey confirmed the hypothesis that the indefinite pronoun is more often used in singular NPs, if no relative clause follows the NP.

In the case when the subclass is anonymous, a reference to the universal class is naturally made (see Table 4) by using an indefinite pronoun. Here the problem of differentiation between animate ('everyone') and inanimate ('everything') things appears (both in English and in Baltic languages). We can easily allow such differentiation from the analysis point of a view, however, the problem is how to choose the appropriate pronoun, if an ontology, which has not been created and annotated by means of a CNL, is being verbalized. To keep the lexicon robust and domain-independent, the ACE verbalizer [4] always uses 'everything'.

**Table 2.** Generalization axiom that refers to an anonymous superclass — in this case, to a disjunction of the named classes. Fragments in square brackets are optional (*vai nu* stands for 'either').

| ACE | | **Every** course **is a** mandatory_course **or is an** optional_course. |
|---|---|---|
| Sg | P | (Ikviens\|Katrs) kurss ir [vai nu] obligātais_kurss vai izvēles_kurss. |
| | L | **Ikviens** kurss **ir** obligātais_kurss **vai** izvēles_kurss. |
| Pl | P | Visi kursi ir [vai nu] obligātie_kursi vai izvēles_kursi. |
| | L | **Visi** kursi **ir** obligātie_kursi **vai** izvēles_kursi. |
| OWL | | Class: Course SubClassOf: MandatoryCourse or OptionalCourse |

---

[4] Manchester OWL Syntax [8], for instance, which is intended to be the most user-friendly formal syntax for OWL. We have used it for comparison with CNL statements in all examples (except Table 7 and Table 10) in this paper.
[5] Note that while we are dealing only with the terminological axioms, this is only a grammatical issue, which does not introduce any interpretation ambiguities: if a NP is not explicitly universally quantified we could assume that it is existentially quantified.

**Table 3.** Generalization axiom that includes a property restriction. The indefinite pronoun *kāds* is optional, but is used in the linearization (in the singular only), as there is no relative clause attached with the NP.

| ACE | | **Every** course **is** taught **by a** teacher. |
|---|---|---|
| Sg | P | (Ikvienu\|Katru) kursu (māca\|pasniedz) [kāds] pasniedzējs. |
| | L | **Ikvienu** kursu$_{ACC}$ māca **kāds** pasniedzējs. |
| Pl | P | Visus kursus (māca\|pasniedz) pasniedzēji. |
| | L | **Visus** kursus$_{ACC}$ māca pasniedzēji. |
| OWL | | Class: Course SubClassOf: inverse (teaches) some Teacher |

In order to linearize the appropriate pronoun, additional information should be encoded for each lexical unit, indicating whether the term represents animate or inanimate things. This would make the lexicon sense- and, thus, domain-specific.

Few Latvian respondents suggested to use the neutral pronoun *tas* ('that') for 'everything'/'everyone'. It is unlikely that a controlled Latvian user would intuitively use it himself, however, if the pronoun is automatically used in linearization, the statement remains grammatically correct and easily comprehensible. In the Lithuanian questionnaire we used its counterpart (*tai*) by default, which was generally accepted.

The issue, however, cannot be avoided in statements defining domain or range of a property (see Table 5). In such definitions both the subject and the object are references to the universal class; one of them — existentially quantified ('something') without any restricting relative clause. There is no neutral pronoun (in all the three languages) that could be used instead of 'something'. Moreover, to include the information in the lexicon, it should be encoded in verb entries, indicating whether the subject/object has to be an animate or an inanimate thing. Therefore the chosen trade-off for linearization is to use the indefinite pronoun *kaut kas* ('something'), which is more neutral.

Many Latvian respondents also suggested to use the personal pronoun *kurš* instead of the relative pronoun *kas*, if the antecedent of the anaphor is an animate thing. Such differentiation is probably influenced from Russian, but due to its frequent use both pronouns are included in the lexicon; for linearization the relative one is always used.

**Table 4.** Generalization axiom that refers to an anonymous subclass The pronoun 'everything' refers to the universal class *owl:Thing*, which is further specified by the property restriction.

| ACE | | **Everything that** teaches **a** mandatory_course **is a** professor. |
|---|---|---|
| Sg | P | (Tas\|Ikviens\|Katrs\|Jebkas\|Viss), (kas\|kurš) (māca\|pasniedz) [kādu] obligāto_kursu, ir profesors. |
| | L | **Tas**, **kas** māca **kādu** obligāto_kursu$_{ACC}$, **ir** profesors. |
| Pl | P | (Tie\|Visi), (kas\|kuri) (māca\|pasniedz) obligātos_kursus, ir profesori. |
| | L | **Tie**, **kas** māca obligātos_kursus$_{ACC}$, **ir** profesori. |
| OWL | | Class: owl:Thing and (teaches some MandatoryCourse) SubClassOf: Professor |

**Table 5.** Axiom defining the range of a property. The domain would be defined in the active voice. Note that the domain/range definitions can be distinguished from generalizations only due to the agreement that both the subject and the object refers to the universal class and the existentially quantified one is not restricted.

| | ACE | **Everything that is** taught **by something is a** course. |
|---|---|---|
| Sg | P | (Tas\|Jebkas\|Viss\|Ikviens\|Katrs), (ko\|kuru) (kaut kas\|kāds) (māca\|pasniedz), ir kurss. |
| | L | **Tas**, **ko**$_{ACC}$ **kaut kas** māca, **ir** kurss. |
| Pl | P | (Tie\|Visi), (ko\|kurus) (kaut kas\|kāds) (māca\|pasniedz), ir kursi. |
| | L | **Tie**, **ko**$_{ACC}$ **kaut kas** māca, **ir** kursi. |
| | OWL | ObjectProperty: teaches Range: Course |

So far we have not faced anaphoric references, except the relative pronouns that start subordinate clauses. Apart from the assertional (factual) statements that are not covered in this paper, the need for anaphors emerges when SWRL rules and SPARQL integrity queries are verbalized (see Table 7 and Table 10 in the next sections). It was already mentioned that in the case of synthetic CNL we can impose the use of systematic word order patterns [6], e.g., if a NP stands before the verb, it should be an anaphor (if it is not universally quantified). The majority of respondents confirmed that the word order changes are intuitive, but the majority also requested that in rules and queries, which typically contain more than one subordinate clause, the definite pronoun should be used as well. Therefore, for the unambiguous parsing the pronoun is still optional (parsing is fully TFA-based), but it is always used in linearization.

## 2. Pseudo-SVO Statements

Theoretically, all predicates in OWL ontologies should conform to the SVO pattern (generalization predicates are a special case). In practice, however, it can be very hard or even impossible to come up with an appropriate verb for a property, or to use a syntactic object (accusative case), so that the statement would remain natural.

In the first case, individuals of two classes most likely can be associated at least by means of a role (a NP) that would be translated into OWL as a property. The leading CNLs support different, but limited ways how to define and refer such properties [1]. In controlled Baltic languages they can be expressed in a uniform way: by making a NP that consists of a class name in the genitive (possessive) case followed by its role name whose case depends on the context (see an example in Latvian in Table 6).

**Table 6.** Use of a noun instead of a verb (predicate) to express an association (property) between two classes.

| | ACE | **Every** course **is a** part **of an** academic_program.<br>**For every** academic_program **its** part **is a** course. |
|---|---|---|
| Sg | P | (Ikviens\|Katrs) kurss ir [kādas] akadēmiskās_programmas daļa.<br>(Ikvienas\|Katras) akadēmiskās_programmas daļa ir [kāds] kurss. |
| | L | **Ikviens** kurss **ir kādas** akadēmiskās_programmas$_{GEN}$ daļa.<br>**Ikvienas** akadēmiskās_programmas$_{GEN}$ daļa **ir kāds** kurss. |
| | OWL | Class: Course SubClassOf: inverse (part) some AcademicProgram<br>Class: AcademicProgram SubClassOf: part some Course |

**Table 7.** Use of a modifier instead of an object: *course-included_in-program* vs. *program-includes-course*.

| ACE | | **Every** student takes **every** mandatory_course **that is** included_in **an** academic_program *that* enrolls **the** student. |
|---|---|---|
| Sg | P | (Ikviens\|Katrs) students (apgūst\|ņem) (ikvienu\|katru) obligāto_kursu, (kas\|kurš) ([ir] iekļauts\|ietilpst) [kādā] akadēmiskajā_programmā, kurā [šis] students [ir] uzņemts. |
|  | L | **Ikviens** students apgūst **ikvienu** obligāto_kursu$_{ACC}$, **kas ir** iekļauts akadēmiskajā_programmā$_{LOC}$, ***kurā*** $_{LOC}$ ***šis*** students **ir** uzņemts. |
| SWRL | | AcademicProgram(?x3), MandatoryCourse(?x2), Student(?x1), enrolls(?x3, ?x1), includes(?x3, ?x2) → takes(?x1, ?x2) |

In the second case, typical pseudo-objects in ontological statements are adverbial modifiers of place. Apart from statements where nothing else than a modifier is possible, the survey confirmed that use of an object is inappropriate also in statements where it is syntactically possible, but semantically incorrect (e.g. 'academic program' in Table 7). Note that currently we are considering only such modifiers that in English require the preposition 'in', but in Baltic languages are expressed by the locative case.

There is an issue, however, in translating relative clauses to/from ACE, if the relative pronoun is the modifier — such constructions are not supported in ACE. The transformation form modifier to object clauses can be easily introduced in the parallel GF grammars, but ambiguity arises in the opposite direction. To ensure the correct choice between object and modifier constructions, morphological restrictions on verb arguments have to be encoded in the lexicon; if both constructions could be valid, they can be included as alternatives, indicating which is preferable for linearization.

Table 7 illustrates one more aspect: although in OWL (and FOL in general) there is no time dimension, the survey showed that majority of CNL users would prefer to differentiate perfect and imperfect actions. To support this, use of certain participles has been allowed at the surface level (in both SVO and pseudo-SVO statements).

### 3. Negated Statements

Three typical cases when negation has to be used are: to define disjoint classes (see Table 8), to define a subclass of the complement of a class (see Table 9), and to ask a data integrity query (see Table 10). In all cases, no determiners are used in the plural.

**Table 8.** Axiom defining disjoint classes. Double negation, in general, is used in both Latvian and Lithuanian.

| ACE | | **No** assistant **is a** professor. |
|---|---|---|
| Sg | P/L | **Neviens** asistents **nav** profesors. |
| OWL | | DisjointClasses: Assistant, Professor |

**Table 9.** Generalization axiom that includes a negated property restriction. The indefinite pronoun *neviens* is optional (in the object position), but is used in the linearization, because no relative clause follows.

| ACE | | **No** assistant teaches **a** mandatory_course. |
|---|---|---|
| Sg | P | Neviens asistents (nemāca\|nepasniedz) [nevienu] obligāto_kursu. |
|  | L | **Neviens** asistents nemāca **nevienu** obligāto_kursu$_{ACC}$. |
| OWL | | Class: Assistant SubClassOf: not (teaches some MandatoryCourse) |

**Table 10.** Data integrity query. In the singular, the indefinite pronoun *kāds* is always used with the subject.

| ACE | | **Is there a** student **that** takes **a** course **that is not** included_in **an** academic_program *that* enrolls **the** *student*? |
|---|---|---|
| Sg | P | Vai ir kāds students, (kas\|kurš) (apgūst\|ņem) [kādu] kursu, (kas\|kurš) (nav iekļauts\|neietilpst) [nevienā] akadēmiskajā_programmā, kurā [šis] students [ir] uzņemts? |
| | L | **Vai ir kāds** students, **kas** apgūst kursu$_{ACC}$, **kas nav** iekļauts akadēmiskajā_programmā$_{LOC}$, **kurā**$_{LOC}$ *šis* **students ir** *uzņemts*? |
| SPARQL | | ASK WHERE {?x1 rdf:type Student. ?x1 takes ?x2. ?x2 rdf:type Course. ?x3 rdf:type AcademicProgram. ?x3 enrolls ?x1. NOT EXISTS {?x3 includes ?x2}} |

## 4. Conclusion

An interesting observation was made that after a little training one can easily express rather complex rules (if they are conceptually clear to him), however, it might not be easy for others to grasp the meaning, if cascades of relative clauses are used. Transforming the relative clauses into genitive NPs often improves the readability — it should be supported as an alternative in future. Other tasks are to extend the grammars to support prepositional phrases as adverbial modifiers, cardinality constraints of properties, *if-then* statements as an alternative pattern, and assertional (ABox) statements.

A trade-off has to be found regarding the lexicon: whether it will be domain-independent or domain-specific. On the one hand, the latter choice enables to differentiate animate/inanimate things, to impose morphological restrictions on verb arguments etc. It would also allow adapting of existing, linguistically non-motivated ontologies for verbalization. On the other hand, GF does not support anaphora resolution — if we intend to use existing tools for translation to/from OWL, it is important to keep compliance with ACE or some other well resourced CNL.

## Acknowledgements

The research is funded by the State Research Programme in Information Technologies. The authors would like to thank Guntis Bārzdiņš for encouraging the research topic, and all respondents for their significant help in evaluation of the proposed grammars.